# A Schedule of Duties in the Cloud Space Using a Modified Salp Swarm Algorithm

Hossein Jamali, Ponkoj Chandra Shill,
David Feil-Seifer, Frederick C. Harris, Jr., Sergiu M. Dascalu
Department of Computer Science and Engineering
University of Nevada, Reno
Reno, NV, USA
{hossein.jamali,ponkoj}@nevada.unr.edu,
{dave,fred.harris,dascalus}@cse.unr.edu

**Abstract.** Cloud computing is a concept introduced in the information technology era, with the main components being the grid, distributed, and valuable computing. The cloud is being developed continuously and, naturally, comes up with many challenges, one of which is scheduling. A schedule or timeline is a mechanism used to optimize the time for performing a duty or set of duties. A scheduling process is accountable for choosing the best resources for performing a duty. The main goal of a scheduling algorithm is to improve the efficiency and quality of the service while at the same time ensuring the acceptability and effectiveness of the targets. The task scheduling problem is one of the most important NP-hard issues in the cloud domain and, so far, many techniques have been proposed as solutions, including using genetic algorithms (GAs), particle swarm optimization, (PSO), and ant colony optimization (ACO).  To address this problem, in this paper one of the collective intelligence algorithms, called the Salp Swarm Algorithm (SSA), has been expanded, improved, and applied. The performance of the proposed algorithm has been compared with that of GAs, PSO, continuous ACO, and the basic SSA. The results show that our algorithm has generally higher performance than the other algorithms. For example, compared to the basic SSA, the proposed method has an average reduction of approximately 21% in makespan.

# 1    INTRODUCTION

Today, modern computing methods have attracted the attention of researchers in many fields such as cloud computing, artificial intelligence, and machine learning by using techniques including artificial neural networks in building air quality prediction models that can estimate the impact of climate change on future summer trends [1]. A computational science algorithm is used in this article to determine the schedule of duties in the cloud.

Cloud computing has brought about the availability of tools that provide extensive computing resources on the internet platform. Users can submit their requests for various resources, such as CPU, memory, disk, and applications, to the cloud provider. The provider then offers the most suitable resources, which meet the user's requirements and offer benefits to the resource owners, based on the price that they can afford to pay [2]. In cloud computing, the main entities are users, resource providers, and a scheduling system whose main body has been proposed for the users' tasks and timeline strategy [3].

Cloud computing consumers rent infrastructure from third-party providers instead of owning it. They opt for this to avoid extra costs. Providers typically use a "pay-as-you-go" model, allowing customers to meet short-term needs without long-term contracts, thus reducing costs. [4].

Behind the numerous benefits of cloud computing, there are many challenges too. The most important is the task scheduling problem or resource allocation to the users' requests. The targets of task scheduling in cloud computing are to provide operating power, the optimal timeline for users, and service quality simultaneously. The specific targets related to scheduling are load balance, service quality, economic principles, the best execution time, and the operating power of the system [5]. Cloud computing has three timelines: resources, workflow, and tasks. Resource scheduling involves mapping virtual resources to physical machines. Workflow scheduling ensures the orderly flow of work. Task scheduling assigns tasks to virtual resources. Task scheduling methods can be concentrated or distributed, homogeneous or heterogeneous, and performed on dependent or independent tasks.

Task scheduling in cloud computing has two types based on the characteristic of the tasks:
- *Static:* In static scheduling, the tasks reach the processor simultaneously and are scheduled on accessible resources. The scheduling decisions are made before reaching the tasks and the processing time after doing the entire run of duty is updated. This type of scheduling is mostly employed for tasks that are sent continuously [6]; and
- *Dynamic:* In dynamic scheduling, the number of tasks, the location of the virtual machines, and the method for resource allocation are not constant, and the input time of tasks before sending them is unknown [6].

Scheduling the mechanism of dynamic algorithms compared to static algorithms is better but the overhead of the dynamic algorithm is quite significant [7]. Dynamic scheduling can be done in two ways; in batch and online modes. In batch mode, the tasks are lying in a line, gathered in a set, and after a certain time, scheduled. In the online mode, when the tasks reach the system, they are scheduled [6].

The task scheduling problem in cloud computing focuses on efficiently distributing tasks among machines to minimize completion time [8]. Proper task arrangement has numerous benefits, including reduced energy consumption, increased productivity, improved distribution across machines, shorter task waiting times, decreased delay penalties, and overall faster task completion [9].

The task scheduler plays a crucial role in efficiently scheduling computing actions and logically allocating computing resources in IaaS cloud computing. Its objective is to assign tasks to the most suitable resources to achieve specific goals. Selecting an appropriate scheduling algorithm is essential to enhance resource productivity while maintaining a high quality of service (QoS). Task scheduling involves optimizing the allocation of subtasks to virtual servers in order to accomplish the task schedule's objective. This area of research continues to receive significant attention [10].

Efficient task planning in cloud computing is essential to minimize fetch time, waiting time, computing time, and resource usage. Task scheduling is crucial for maximizing cloud productivity, meeting user needs, and enhancing overall performance. Its primary goal is to manage and prioritize tasks, reducing time and preventing work failures while meeting deadlines. Task scheduling optimizes the cloud computing system for improved calculation benefits, high performance, and optimal machine output. The scheduling algorithm distributes work among processors to maximize efficiency and minimize workflow time [11].

The rest of this paper is organized as follows: Section 2 covers related work; Section 3 provides details of the SDSA optimization algorithm; Section 4 describes our proposed method, including the expansion and improvement of the salp algorithm; Section 5 focuses on the algorithm's target, the fitness function; Section 6 presents the results of our simulation; and, finally, Section 7 contains the conclusions of our work.

## 2 RELATED WORKS

Ghazipour et al. [12] have proposed a task scheduling algorithm so the tasks existing in the grid are allocated to accessible resources. This algorithm is based on the ACO algorithm, which is mixed with the scheduling algorithm right to choose so that its results are used in the ACO algorithm. The main goal of their article is to minimize the total finish time (makespan) for setting up tasks that have been given [12].

In their research on task scheduling in cloud computing, Sharma and Tyagi [13] examined nine heuristic algorithms. They conducted comparative analyses based on scheduling parameters, simulation tools, observation domain, and limitations. The results indicated the existence of a heuristic approach that satisfies all the required parameters. However, considering specific parameters such as waiting time, resource utilization, or makespan for each task or workflow individually can lead to improved performance. [13].

In 2019, Mapetu et al. [14] researched the "binary PSO algorithm for scheduling the tasks and load power in cloud computing". They introduced a binary version of the PSO algorithm named BPSO with lower complexity and cost for scheduling the tasks and load power in cloud computing, to minimize waiting time, and imbalance degree while minimizing resource use. The results showed that the proposed algorithm presents greater task scheduling and load power than existing heuristic algorithms [14].

Saeedi et al. [15] studied the development of the multi-target model of PSO for scheduling the workflow in the cloud areas. They proposed an approach for solving the scheduling problem considering four contrasting goals (i.e., minimizing the cost, waiting time, energy consumption, and maximizing reliability). The results showed that the proposed approach had a better performance compared to LEAF and EMS-C algorithms [15].

Zubair et al. [10] presented an optimal task scheduling method using the modified symbiotic organisms search algorithm (G_SOS) and aimed to minimize the makespan of the tasks, costs, response time, and imbalance degree, and improve the convergence speed. The performance of the proposed method using CloudSim (a simulator tool) was evaluated and according to the simulation results, the proposed technique has better performance than the SOS and PSO-Simulated Annealing (PSO-SA) in terms of the convergence speed, cost, imbalance degree, response time, and makespan. The findings confirm the suggested G_SOS approach [10].

Rajagopalan et al. [16] introduced an optimal task-scheduling method that combines the firefly optimization algorithm with a genetics-based evolutionary algorithm. This hybrid algorithm creates a powerful collective intelligence search algorithm. The proposed method excels in minimizing the makespan for all tasks and quickly converges to near-optimal solutions. The results demonstrated that this hybrid algorithm outperformed traditional algorithms like First in, First Out (FIFO) and genetics. However, a potential drawback of this method is the increased overload resulting from the sequential use of two algorithms [16].

## 3 SSA OPTIMIZATION ALGORITHM

This section briefly describes the SSA optimization algorithm proposed by Mirjalini Al which is an extension of the standard SSA algorithm [17]. salps are a type of Salpidae family and have a transparent and barrel-shaped body. Their bodies are very similar to jellyfish. They still move the same as jellyfish, and water is pumped from the middle of the body as a motive force to move forward [17]. The shape of salp is shown in Figure 1(a).

The biological study of these animals is just starting because it is so difficult to capture them and maintain them in laboratories. One of the most intriguing habits of salps is their tendency to swarm. The salps commonly form a chain in the deep oceans. This chain is shown in Figure 1(b). Although the primary cause of this behavior is unknown, some researchers think that it is carried out through quick coordinated movements and searches to achieve improved movement [17].

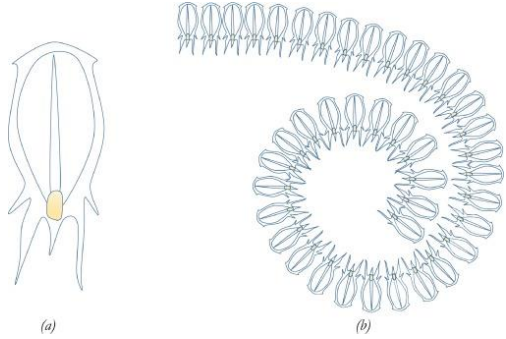

**Fig. 1.** (A) Illustration of a salp. (B) Salp chain structure. [17].

To model mathematically the salp chains, first, the population is divided into two groups: leaders and followers. The leader is in front of the chain, while the remaining are considered the followers. As seen from their names, the leader guides the group and the followers follow each other [17].

Like other techniques based on the swarm, the location of salps in a search space is n-dimensional, where n is the number of variables in a problem and known; therefore, the location of all salps is stored in the two-dimensional matrix x. Also, it is assumed that a food source, F, exists in the search space as a swarm target [17].

Equation 1 has been proposed for updating the location of the leader as follows:

$$x_j^1 = \begin{cases} F_j + c_1\left((ub_j - lb_j)c_2 + lb_j\right) & c_3 \geq 0 \\ F_j - c_1\left((ub_j - lb_j)c_2 + lb_j\right) & c_3 < 0 \end{cases} \quad (1)$$

Where $x_j^1$ shows the location of the first salp (leader) in the $j$ dimension, $F_j$ is the location of the food source in the $j$ dimension, $ub_j$ identifies the upper boundary of the $j$ dimension, $lb_j$ identifies the lower boundary of the $j$ dimension, and $c_1$, $c_2$, and $c_3$ are random numbers (between 0,1) [17].

Equation 1 shows that the leader just updates its location according to the food source. The $c_1$ the constant is the most important parameter in the SSA because it creates a balance between exploration and detection and is defined as equation 2:

$$c_1 = 2e^{-\left(\frac{4l}{L}\right)^2} \quad (2)$$

Here, $l$ is the current iteration and $L$ is the maximum iteration.

The parameters of $c_2$ and $c_3$ are the random numbers which are uniformly produced in the range [0.1]. They determine if the latter location in the $j$ dimension should be infinite positive or infinite negative, as well as determine the step size.

To update the followers' location, equation 3 is used (Newton's law of motion):

$$x_j^i = \frac{1}{2}at^2 + v_0 t \quad (3)$$

If $i \geq 2$, $x_j^i$ shows a salp follows the $i$ location in the $j$ dimension, $t$ is the time, and $v_0$ is the initial velocity; $a = \frac{v_{final}}{v_0}$ and $v = \frac{x-x_0}{t}$. Since the time is iterated in the optimization, the difference between the iterations is equal to 1 and, considering $v_0 = 0$, this relation is expressed as equation 4.

$$x_j^i = \frac{1}{2}\left(x_j^i + x_j^{i-1}\right) \tag{4}$$

Here, $i \geq 2$ and $x_j^i$ shows a salp follows the $i$ location in the $j$ dimension.

The salp chains can be simulated by equations 1 and 4. In the SSA model, the followers follow the salp leader. The leader also moves towards the food source; therefore, if the food source is substituted for the global optimization, the salp chain automatically moves towards it. However, there is a problem that global optimization is unknown in the optimization problems. In this way, it is assumed that the best solution obtained so far is the global optimum, which is assumed as a food source for following the salp chain.

The pseudo-code for the SSA algorithm is shown in Figure 2 [17]. This figure shows that the SSA algorithm begins the global optimum by starting several salps at random locations. Then, each fitting related to the salps are calculated and the location where they have acquired the best fitting is allocated to the variable $F$ as a food source followed by the salp chain. Meanwhile, the value of $c_1$ constant is updated by equation 2. For every dimension, the location of the leader is updated by relation 1 and that of the followers by equation 4. If each salp goes out of the search space, they are returned to the border again. All the mentioned stages except for the initial value are iterated till consent is obtained.

The computing complexity of the SSA algorithm is considered as $O(t(d * n + Cof * n))$ where $t$ shows the number of iterations, $d$ is that of variables (dimension), $n$ is that of solutions, and $Cof$ is a target cost of the function.

```
Initializes the salp population x_i (i = 1,2, ..., n) considering ub and lb
While (end condition is not satisfied)
Calculate the fitness of each search agent(salp)
F= the best search agent(salp)
Update c_1 by Eq. (2)
    for each salp (x_i)
        If (i==1)
            Update the position of the leading salp by Eq. (1)
        else
            Update the position of the, follower salp by Eq. (4)
        end
    end
    Amend the salps based on the upper and lower bounds of variables.
end
returnF
```

**Fig. 2.** Pseudo-code of the salp swarm algorithm. [17].

## 4   PROPOSED METHOD

Our proposed method for scheduling the tasks of the virtual machines in the cloud computing area uses an optimized SSA based on the fitness function. First, a set of random answers created is assigned as the initial population. Each member of this set is called a salp. In the first stage, the fitness of salps produced randomly is calculated by the target function and the best slap is chosen among all salps and its location is determined by the location of the food source. In the following, the salps move towards the food source until they achieve the best food source (i.e., solution). In this algorithm, each salp is represented as a solution that moves for searching based on a mechanism in the problem space. In the suggested method, the salps are divided into two groups, the leaders and the followers. One group of salps named leader salps updates its location according to the food source and tries to move towards the existing food source and discover a better solution. If they find a better solution than the existing food source, the location of the leader salp is considered as its new location. The group salps follow each other, and if they discover a better solution for the food source location, the location of the salp follower is considered the new location of the food source.

### 4.1   The task scheduling problem in the cloud area

The task scheduling problem in the cloud is allocating the settings of tasks to a set of sources. We have assumed a set of $n$ tasks, $T = (T_1.T_2.T_3.\cdots.T_n)$, and of $m$ sources, which are virtual machines in targeted source research, $V = (V_1.V_2.V_3.\cdots.V_m)$. The set of $T$ includes the tasks which should be scheduled. Each task should be processed by virtual machines so that the completion time of all tasks is minimized as much as possible.

The main goal of task scheduling is to allocate optimally to the sources so that the lowest completion time of the tasks (i.e., makespan) and the lowest cost is obtained. The makespan shows the total required time for implementing all the tasks. The main goal of our research is to minimize the makespan using the modified SSA.

### 4.2   The proposed coding method

Assume that an array of 200 tasks exists and each task has a value between 1-15. For example, if the second value of this array is 4, it shows that task 2 has been allocated to the virtual machine 4 and, if the seventh value of the array is 14, it means that, task 7 has been allocated to the virtual machine 14. Similarly, all the tasks $T_1$ to $T_{200}$ are allocated to virtual machines $V_1$-$V_{15}$. In Figure 3, an example of allocating tasks to virtual machines is depicted.

| $T_1$ | $T_2$ | $T_3$ | ... | $T_i$ | ... | $T_{200}$ |
|---|---|---|---|---|---|---|
| $V_2$ | $V_4$ | $V_{14}$ | ... | $V_1$ | ... | $V_7$ |

**Fig. 3.** Allocation of tasks to virtual machines.

In the suggested algorithm, solutions are shown by a salp chains. Each solution of the suggested algorithm is shown by an array of natural numbers. The locations of all salps are stored in a 2-dimensional matrix named $x$. For instance, in a problem with $n$ tasks and $m$ virtual machines, the rows of a two-dimension matrix are considered as the number of the salp population. It means that the location of each salp is restored in a row of a matrix. The columns of the matrix are equal to $n$. Also, the content of each cell of the array shows the virtual machine number, which can be a number between 1 to $m$. Figure 4 shows an example of a salp.

| $T_1$ | $T_2$ | $T_3$ | $T_4$ | $T_5$ | $T_6$ | $T_7$ | $T_8$ |
| ↓ | ↓ | ↓ | ↓ | ↓ | ↓ | ↓ | ↓ |
| 4 | 2 | 3 | 4 | 5 | 2 | 1 | 3 |

**Fig. 4.** An example of a salp.

To begin the work, this salp can be produced as a desired number where this number is the same as the primary population of the algorithm that is adjusted. First, the population is randomly generated and stored in a two-dimensional matrix where its rows are identical to the number of salps and its columns equal to those of tasks identified for the scheduling.

After generating the primary population of salps in the range of the problem answer, the fitness of all salps is assessed by all salps and the salp with the best fitness is determined. In this algorithm, it is assumed that a food source named $F$ exits in a search space as a swarm target that all salps try to move towards it.

In the first stage of this algorithm, the location of the best generated salp (the best solution) is considered as the food source.

In the next stage of this algorithm, the salps are divided into two groups of leaders and followers. The number of salps is considered as the leader salp group and the remaining as the follower one. In the proposed algorithm, 50% of salps are considered as the leader group and the remaining 50% as followers. The location of the leader group is updated by equation 5.

$$x_j^i = F_j + \alpha Randn(\ ) \qquad (5)$$

Where $x_j^i$ is the location of the leader salp $i$, $F_j$ the location of the food source in the $j$ dimension, $\alpha$ the constant of the random moving step in the range of [0,1] that is adjusted by the targeted problem, and $Randn(\ )$ a random number with a normal distribution and determines a random step with a normal distribution for the leader group. Equation 6 updates the location of the follower group.

$$x_j^i = \frac{1}{2}\left(x_j^i + x_j^{i-1}\right) + c_1 Randn(\ ) \qquad (6)$$

Where $x_j^i$ is the location of the follower salp $i$ in the $j$ dimension. The constant $c_1$ creates a balance between the exploration and discovery by generating an adaptive step, and this constant decrease consistently during the iterations; so, it leads to higher discovery in the first iterations and higher exploration in the end iterations if the algorithm, $Randn(\ )$ is a random number with a normal distribution and determines a random step with this distribution for the leader group. The parameter $c_1$ is defined in equation 7 and is updated in each iteration.

$$c_1 = 2e^{-\left(\frac{4l}{L}\right)^2} \tag{7}$$

Here, $l$ is the current iteration and $L$ the maximum of iterations.

In each iteration of the algorithm, after updating, first, the location of each salp is explored; if each salp goes out of the search space, it returns to the borders. Next, its fitness has been assessed based on the target function; if its fitness has been better than that of the food source, the location of the desired salp has been substituted for that of the food source. It is noted that in the substitution of the salp location for the food source, there is a difference between the leader group and the follower group when swapping. In the case of the leader group, even if the fitness of the leader salp and food source are identical, the location of the leader salp is substituted for the food source, because the salps with equal fitness have different locations, and this mechanism is an effective alternative for diversifying a search space, releasing from the local optimum, as well as discovering accurately surrounding the existing food source.

Based on this, the population of the leader group updated its location using the location of the food source. When the location of each leader salp group is substituted for that of the food source, the latter group has updated its location using the new location of the food source. Figure 5 depicts the algorithm's pseudo-code of the optimized SSA.

The stages of the algorithm until reaching the end are continued. In the proposed algorithm, the condition for finishing the algorithm is the number of iterations.

```
Initializes the salp population xi(i = 1,2, ... , n) considering ub and lb
Calculate the fitness of each search agent(salp) from the fitness function.
 F= the best search agent(salp)
Initialize α
    While (end condition is not satisfied)
        Update c1 by Eq. (7)
            For each salp (xi)
                If (i<=N* 0.5)
                Update the position of the leading salp by Eq. (5)
                Amend the sales based on the upper and lower bounds of variables.
                Calculate the fitness of the leading salp from the fitness function.
                    If (the fitness of the leading salp <= the fitness of the F)
                    F= position of the leading salp
                    End If
                else
                Update the position of the follower salp by Eq. (6)
                Amend the salps based on the upper and lower bounds of variables.
                Calculate the fitness of the follower salp from the fitness function.
                    If (the fitness of the follower salp < the fitness of the F)
                    F= position of the follower salp
                    End If
                End If
            End For
        End While
returnF
```

**Fig. 5.** The pseudo-code of the modified SSA.

## 5 FITNESS FUNCTION

The main goal of this research is to minimize the makespan, one of the most important targets for the task scheduling problem in the cloud areas. An example of task samples and task sizes is given in Table 1 and another is shown in Table 2 for virtual machines and the processor speed as individual values.

Table 1. An example of the tasks and their size.

| Tasks | $T_1$ | $T_2$ | $T_3$ | $T_4$ | $T_5$ | $T_6$ | $T_7$ | $T_8$ | $T_9$ | $T_{10}$ | $T_{11}$ | $T_{12}$ |
|---|---|---|---|---|---|---|---|---|---|---|---|---|
| Size | 18 | 15 | 19 | 24 | 33 | 41 | 22 | 12 | 30 | 16 | 13 | 32 |

Table 2. An example of virtual machines and their speed.

| Virtual machine number | 1 | 2 | 3 | 4 | 5 |
|---|---|---|---|---|---|
| Processor speed | 3.4 | 2.4 | 3.2 | 1.8 | 2.2 |

We aim to reduce the completion time of tasks in this research. This time duration is the longest completion time among virtual machines. If we consider $T_i$ as the task size of $i$ and $C_j$ as the processor speed of the virtual machine $j$, we can obtain the makespan $i$ from equation 8.

$$t_{exe}(i.j) = T_i/C_j \qquad (8)$$

According to the allocated tasks for each resource and the length of desired tasks, there has been a completion time for tasks relative to the processor speed of the virtual machine for each of them.

For instance, assume that the tasks $T_3, T_6, T_{10}, T_8$ are allocated to virtual machine 2, the makespan of each task delivered to virtual machine 2 can be calculated as follows:

$$t_{exe}(3.2) = \frac{19}{2.4} = 7.9 \qquad t_{exe}(6.2) = \frac{41}{2.4} = 17.1$$
$$t_{exe}(10.2) = \frac{16}{2.4} = 6.7 \qquad t_{exe}(8.2) = \frac{12}{2.4} = 5$$

So, the completion time of tasks calculated on virtual machine 2 is:

$$t_{complete}(2) = 7.9 + 17.1 + 6.7 + 5 = 36.7$$

Similarly, the times for all virtual machines can be computed from the assigned tasks. The longest completion time of tasks amongst that for all virtual machines is calculated by equation 9:

$$Makespan = Max\{t_{complete}(j)\} 1 \leq j \leq m \tag{9}$$

In equation 9, $t_{complete}(j)$ shows the completion time of tasks allocated to the virtual machine $j$. Minimizing equation 9 (i.e., the completion time of all tasks (makespan)) is the main target of this research.

## 6   SIMULATION AND RESULTS

In this section, the performance of the proposed algorithm (Modified salp Swarm Algorithm) is evaluated for solving the task scheduling problem in the cloud area and compared with other algorithms such as Standard salp Swarm Algorithm (SSA), Ant Colony Optimization (ACOr), Particle Swarm Optimization (PSO), and Genetic Algorithm (GA) in multiple scenarios[17]. MATLAB software has been used for simulation. The parameters and their initial values of the compared algorithms have been given in Table 3 and their description in Table 4. The simulation was run for four scenarios with parameters shown in Table 5 and the findings of each scenario are depicted in Figure 6 using associated the chart.

**Table 3.** Parameters and the initial values of the compared algorithms.

| Algorithm | Parameters and the initial values of the algorithms |
|---|---|
| GA | nPop=40, MaxIt=500, pc=0.8, pm=0.3, mu=0.02, nc=32, nm=12, beta=8, RWS=0 |
| PSO | nPop=40, MaxIt=500, C1=2, C2=2, w=0.7 |
| ACO | nPop=40, MaxIt=500, nSample=40, q=0.9, zeta=0.1 |
| SSA | nPop=40, MaxIt=500 |
| Modified SSA | nPop=40, MaxIt=500, α=0.19 |

**Table 4.** description of parameters used for comparing the algorithms.

| | |
|---|---|
| For all | MaxIt=Maximum Number of Iterations nPop=Population Size |
| GA | Pc = Crossover Percentage nc = Number of Offsprings (Parnets) pm = Mutation Percentage nm = Number of Mutants mu = Mutation Rate beta =Roulette Wheel Selection (RWS) Pressure RWS = 0 or 1 |
| PSO | c1 = Personal Learning Coefficient w = Inertia Weight c2 = Global Learning Coefficient |
| ACOR | nSample = Archive Size, q=Intensification Factor (Selection Pressure) zeta=Deviation-Distance Ratio |
| Modified SSA | α= Random step coefficient |

Table 5. parameters of the scenarios.

| Scenario | The number of virtual machines | The number of tasks |
|---|---|---|
| First | 10 | 150-200-250-300 |
| Second | 15 | 150-200-250-300 |
| Third | 20 | 150-200-250-300 |
| Fourth | 25 | 150-200-250-300 |

In the experiments, all algorithms used a number of 40 primary populations and a maximum of 500 iterations. Each scenario was run 20 times to obtain the results. The primary objective was to examine and minimize the makespan measure across different scenarios.

Table 6. data results of the first scenario.

| No. of Tasks<br>Algorithm | 300 | 250 | 200 | 150 |
|---|---|---|---|---|
| SSA | 308.00 | 258.34 | 212.50 | 156.15 |
| ACOr | 282.69 | 236.85 | 192.44 | 144.69 |
| PSO | 275.05 | 230.64 | 186.71 | 139.91 |
| GA | 271.71 | 226.82 | 184.80 | 138.48 |
| Average | 284.36 | 238.16 | 194.61 | 144.80 |
| STD | 16.4148 | 14.07 | 12.68 | 8.014 |
| MSSA | 269.80 | 225.39 | 182.41 | 136.09 |
| Average Improvement in MSSA | 5.40% | 5.66% | 6.68% | 6.40% |

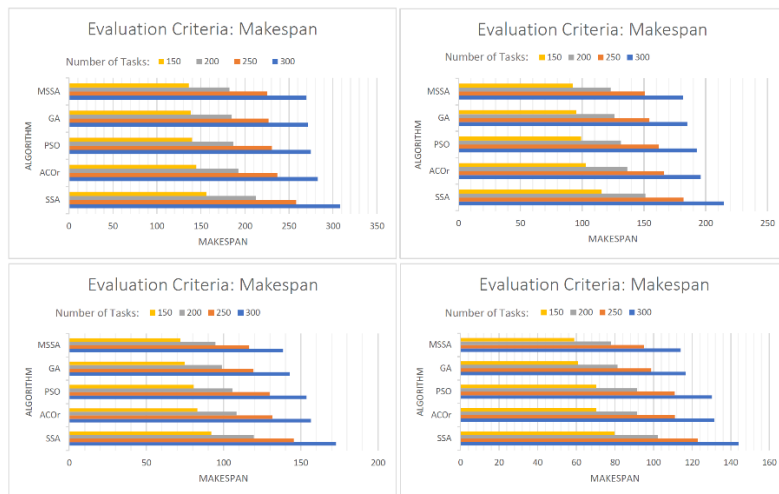

**Fig. 6.** Performance output for the four scenarios, comparing MSSA with other algorithms; MSSA shows lower calculation amount, which is desirable as lower values indicate improved efficiency in minimizing makespan for cloud computing task scheduling.

The results of our simulation study using an Modified Salp Swarm Algorithm (MSSA) for scheduling cloud computing tasks have been analyzed and compared with other well-known optimization algorithms, specifically the Standard salp Swarm Algorithm (SSA), (ACOr), (PSO), and (GA). The simulation results demonstrate that the proposed MSSA algorithm outperforms other algorithms in terms of task completion time.

As shown in Table 6, the MSSA algorithm achieved an average completion time that was 5.40%, 5.66%, 6.68%, and 6.40% better than the average completion time of SSA, ACOr, PSO, and GA, respectively. Furthermore, the standard deviation of the MSSA algorithm was lower than that of other algorithms, indicating more consistent performance. The findings of this study provide valuable insights into the efficiency of different optimization algorithms for scheduling cloud computing tasks. The MSSA algorithm has shown substantial potential in reducing task completion time and improving the overall performance of cloud computing systems. Therefore, it can be concluded that the MSSA algorithm can be a useful tool for scheduling cloud computing tasks in real-world scenarios.

## 7   CONCLUSION

The results from the stated scenarios show that the proposed algorithm had better performance compared to the other algorithms to solve the task scheduling problem in all four scenarios of cloud computing.

The results show that the makespan is reduced by increasing the number of virtual machines and vice versa. They also indicate that the optimized salp swarm algorithm has increased performance compared to the basic one. The outputs of all scenarios were similar and the MSSA is better in all case. As a result, the suggested method has shown better performance in all scenarios to solve the task scheduling problem in the cloud computing domain.

In addition, the findings of this study provide valuable insights into the efficiency of different optimization algorithms for scheduling cloud computing tasks. The MSSA algorithm has shown substantial potential in reducing task completion time and improving the overall performance of cloud computing systems. Therefore, it can be concluded that the MSSA algorithm can be a useful tool for scheduling cloud computing tasks in real-world scenarios.

Overall, while the study's results demonstrate the effectiveness of the MSSA algorithm in reducing task completion time and improving the overall performance of cloud computing systems, it is important to consider the limitations and scope of the study's findings. Future work could explore alternative performance metrics, evaluate the algorithm's robustness and scalability, and investigate its suitability for different cloud computing scenarios.

## Acknowledgment

This material is based in part upon work supported by the National Science Foundation under grant #DUE-2142360. Any opinions, findings, and conclusions or recommendations expressed in this material are those of the authors and do not necessarily reflect the views of the National Science Foundation.